\documentclass[conference]{IEEEtran}
\IEEEoverridecommandlockouts
\usepackage{cite}
\usepackage{amsmath,amssymb,amsfonts}
\usepackage{algorithmic}
\usepackage{graphicx}
\usepackage{textcomp}
\usepackage{xcolor}
\usepackage{booktabs}
\usepackage{lipsum} 
\usepackage{booktabs} 
\usepackage{array}    
\begin{document}

\title{Transforming Game Play: A Comparative Study of DCQN and DTQN Architectures in Reinforcement Learning\\

\thanks{Identify applicable funding agency here. If none, delete this.}
}

\author{\IEEEauthorblockN{1\textsuperscript{st} William A. Stigall}
\IEEEauthorblockA{\textit{College of Computing and Software Engineering} \\
\textit{Kennesaw State University}\\
Marietta, United States of America \\
wstigall@students.kennesaw.edu}
}

\maketitle

\begin{abstract}
In this study we investigate the performance of Deep Q-Networks utilizing Convolutional Neural Networks (CNNs) and Transformer architectures across 3 different Atari Games. The advent of DQNs have significantly advanced Reinforcement Learning, enabling agents to directly learn optimal policy from high dimensional sensory inputs from pixel or RAM data. While CNN based DQNs have been extensively studied and deployed in various domains Transformer based DQNs are relatively unexplored. Our research aims to fill this gap by benchmarking the performance of both DCQNs and DTQNs across the Atari games' Asteroids, SpaceInvaders and Centipede. We find that in the 35-40 million parameter range, the DCQN outperforms the DTQN in speed across both ViT and Projection Architectures, We also find the DCQN outperforms the DTQN in all games except for centipede. 
\end{abstract}

\begin{IEEEkeywords}
component, formatting, style, styling, insert
\end{IEEEkeywords}

\section{Introduction}
The field of Reinforcement Learning (RL) has witnessed substantial progress with the integration of deep learning techniques, particularly through the development of Deep Q-Networks (DQNs), which have revolutionized the way that Agents learn from and interact with their environments. Google Deepmind introduced the Deep Q-Network in 2013 as a means to learn control policies from high-dimensional sensory input in the form of Atari games\cite{mnih2013playing}. This Deep Q-Network achieved human performance across 49 Atari games using a CNN architecture introduced in AlexNet in 2012\cite{Krizhevsky2012ImageNet}.
Our research utilizes the Arcade Learning Environment (ALE) framework and OpenAI Gym to access the ROMS for Atari 2600 games. OpenAI Gym provides a standardized easy-to-use interface for interacting with these game environments\cite{bellemare13arcade,machado18arcade}. Although Transformers have been introduced to RL, they typically have not been used in isolation and are often combined with convolutional or recurrent structures\cite{esslinger2022deep,chen2021decision}. Our study aims to extend this body of work by specifically evaluating the performance of DQNs that employ a Vision Transformer architecture as well as ones in the context of Atari games, thus contributing to the understanding of Transformer-based DQNs without reliance on convolutions or recurrence. We find it difficult to mitigate information loss in Transformers in the 35-40 million parameter range, and our DTQNs were 2 to 3 times slower than their Convolutional counterparts. 

\section{Related Work}
In this section, we will provide an overview of the existing literature on Deep RL and how it's been used to train agents.
\subsection{Deep Q-Networks (DQNs)}
The beginnings of Deep RL can be largely attributed to the introduction of DQNs in the seminal work by Google Deepmind. The DQN by leveraging convolutional neural networks allowed to learn control policies directly from high-dimensional sensory input and the output of the network being the estimate of future rewards. The preprocessing for the algorithm utilizes frame stacking to provide temporal information for the network. Deepmind's DQN was able to surpass human experts in Beam Rider, Breakout, and Space Invaders\cite{mnih2013playing}.
This was built upon in ``Human-level control through deep reinforcement learning'', The DQN was evaluated largely, outperforming both the best linear agents, as well as professional human games testers across 49 games, marking it as the first artificial agent capable of learning and excelling at a diverse set of challenging tasks\cite{Mnih2015HumanlevelCT}. 
\subsection{Advancements in DQN}
There have been numerous improvements tot he original DQN architecture such as Double Q-Learning, a technique designed to mitigate overestimation bias by employing two separate estimators for action values\cite{vanhasselt2015deep}. Dueling, an architecture that decouples the value and advantage functions improving the efficiency of learning and improving training stability\cite{wang2016dueling}. Additionally, Rainbow networks combining elements such as prioritized experience replay, distributional reinforcement learning, multi-step learning and noisy networks\cite{schaul2016prioritized}.
\subsection{Transformers in Reinforcement Learning}
Transformers introcued in the seminal work ``Attention is All You Need'', revolutionized the field of natural language processing by offering an architecture that exclusively relies on self-attention mechanisms, omitting the need for recurrent layers. This architecture captures dependencies regardless of distance in the input sequence\cite{vaswani2023attention}.
While Transformers where first found to make significant gains in natural language processing, they then had exceptional applications in computer vision demonstrating that Transformers could compete with CNNs in image recognition tasks.
In reinforcement learning, Transformers were found to be difficult to optimize in RL objectives, the solution to this was introducing a Gated Transformer which could surpass LSTMs on the DMLab-30 benchmarks\cite{parisotto2019stabilizing}. The Decision Transformer (Chen, R.T.Q., et al. 2019). Decision Transformers offer a unique perspective on RL tasks by framing them as conditional sequence modeling problems. By conditioning an autoregressive model on desired rewards, past states, and actions, the Decision Transformer effectively generates future actions that optimize returns\cite{chen2021decision}.
\subsection{Vision Transformers and Patch Embedding}
Vision Transformers (ViTs) represent a significant shift in how images are processed for recognition tasks. Unlike CNNs, which extract local features through convolutional layers, ViTs treat an image as a sequence of patches. These patches are embedded into vectors and processed by the Transformer architecture, enabling the model to capture global dependencies across the image. This approach has shown promising results, with ViTs achieving competitive performance against traditional CNN architectures when trained on large datasets. The key to ViT's effectiveness lies in its patch embedding technique, which converts image patches into a format that the Transformer can process, allowing for a more flexible and efficient way to understand both local and global image features\cite{dosovitskiy2021image}.
\subsection{Reinforcement Learning Techniques}
Experience Replay is a critical technique in the training of Deep Q-Networks (DQNs), enhancing both learning stability and efficiency. Initially introduced by Lin in 1992, the concept involves storing the agent's experiences at each timestep in a replay buffer and then randomly sampling mini-batches from this buffer to perform learning updates. This method addresses two main issues in reinforcement learning: the correlation between consecutive samples and the non-stationary distribution of data[reference]. The use of separate target and policy networks is a technique introduced to stabilize the training of DQNs. This approach, detailed in the work by Mnih et al. in 2015, involves maintaining two neural networks: the policy network, which is updated at every learning step, and the target network, which is updated less frequently
The policy network is used to select actions, while the target network is used to generate the target Q-values for the learning updates. This separation helps to mitigate the issue of moving targets in the learning process, where updates to the Q-values could otherwise lead to significant oscillations or divergence in the learning process. By keeping the target Q-values relatively stable between updates, the training process becomes more stable and converges more reliably[reference]. Huber loss is a loss function introduced by Huber in 1964, which has found application in DQNs due to its advantages over other loss functions like the mean squared error (MSE) Huber loss is less sensitive to outliers in data than MSE, as it behaves like MSE for small errors and like mean absolute error (MAE) for large errors. This property makes it particularly useful in the context of DQNs, where the distribution of the temporal-difference (TD) error can have large outliers. By using Huber loss, DQNs can achieve more robust learning, as the loss function prevents large errors from disproportionately affecting the update of the network weights. This contributes to the overall stability and efficiency of the learning process in DQNs[reference].

\section{Methodology}
Describe your research methodology. Detail the procedures for data collection and analysis. 
\begin{figure}[ht]
    \centering
    \begin{minipage}[b]{0.15\textwidth}
        \centering
        \includegraphics[width=\textwidth]{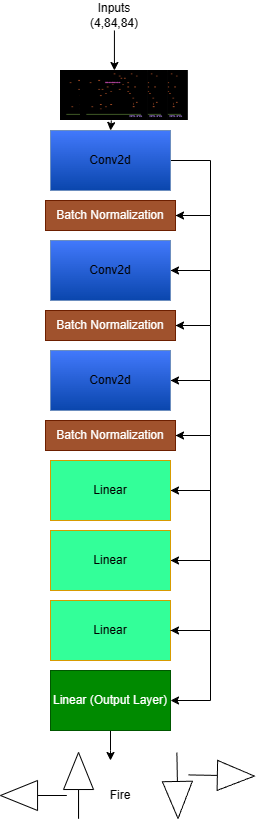}
        \caption{Deep Q-Network architecture, emulates the traditional Q-Network defined in\cite{mnih2013playing}}
        \label{fig:figure1}
    \end{minipage}
    \hfill 
    \begin{minipage}[b]{0.15\textwidth}
        \centering
        \includegraphics[width=\textwidth]{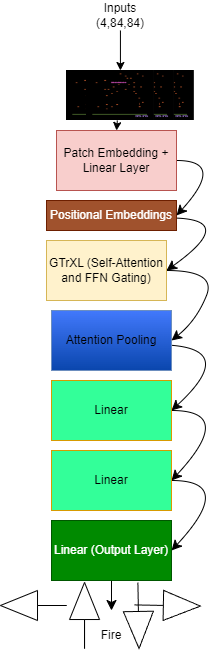}
        \caption{ViT inspired Deep Transformer Q-Network Architecture, utilizes 16x16 patches as the input for the GTrXL.}
        \label{fig:figure2}
    \end{minipage}
    \hfill 
    \begin{minipage}[b]{0.15\textwidth}
        \centering
        \includegraphics[width=\textwidth]{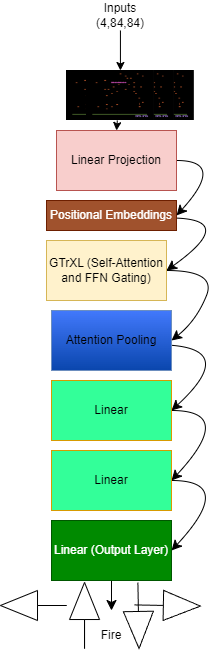}
        \caption{Linear Projection Deep Transformer Q-Network Architecture, increases computational efficiency by opting to project features to the embedding dimension directly.}
        \label{fig:figure3}
    \end{minipage}
\end{figure}
\subsection{Enviornments}
To streamline the learning process and enhance the efficiency of our RL agents we reduce the action space using the specifications provided by OpenAI's gym library for each game. This mitigates the complexity of the decision-making process.
To ensure consistency and fairness across the comparative analysis between the DCQN and DTQN models, we use a fixed seed of 42 for the first environment of the training run, ensuring that all agents are exposed to an identical initial condition.
Frame Stacking is implemented to create an explicit temporal dimension for Positional Embedding for the Transformer model. \subsubsection{State Preprocessing} For preparation for DQN training, we convert each frame from the environment to grayscale. The frames are resized to 110x84 and then center-cropped to 84x84 pixels before being normalized with a mean of 0.5 and a standard deviation of 0.5.
\subsection{DCQN Architecture}
\textbf{Input Representation:}
Let \( I \in \mathbb{R}^{B \times F \times H \times W} \) represent the input batch, where \( B \) is the batch size, \( F \) is the number of stacked frames (acting as channels), and \( H \) and \( W \) are the height and width of the frames, respectively. Each frame is an 84x84 grayscale image, so \( H = 84 \) and \( W = 84 \). The DCQN model processes \( I \) through a series of operations as follows:

\textbf{Convolutional Layers:} The input is passed through three convolutional layers. Each convolutional operation, denoted by \( C_i \), includes a convolution with ReLU activation followed by batch normalization:
\begin{align*}
X_1 &= \text{ReLU}\left(\text{BN}\left(C_1(I)\right)\right) \\
X_2 &= \text{ReLU}\left(\text{BN}\left(C_2(X_1)\right)\right) \\
X_3 &= \text{ReLU}\left(\text{BN}\left(C_3(X_2)\right)\right)
\end{align*}
where \( C_i \) represents the \( i \)-th convolutional layer. The kernel sizes and strides are implicit in \( C_i \) but are not specified here to keep the representation general.

\textbf{Flattening:} The output of the last convolutional layer \( X_3 \) is flattened to form a vector \( X_{\text{flat}} \).

\textbf{Fully Connected Layers:} The flattened vector \( X_{\text{flat}} \) is then passed through three fully connected layers with ReLU activations for the first two layers:
\begin{align*}
X_4 &= \text{ReLU}\left(F_1(X_{\text{flat}})\right) \\
X_5 &= \text{ReLU}\left(F_2(X_4)\right) \\
X_6 &= F_3(X_5)
\end{align*}
where \( F_i \) denotes the \( i \)-th fully connected layer operation. The final output \( X_6 \) represents the Q-values for each action, thus \( X_6 \in \mathbb{R}^{N_{\text{actions}}} \), where \( N_{\text{actions}} \) is the environment-dependent number of possible actions.
\subsection{DTQN Architecture}
The DTQN model is designed to leverage the Transformer architecture for processing sequences of stacked frames from the environment. It consists of several key components: dimensionality reduction, positional embeddings, a Transformer encoder, and fully connected layers for action value prediction. The model can be formalized as follows:

\textbf{Input Representation:}
Let \( X \in \mathbb{R}^{B \times (F \cdot 84 \cdot 84)} \) represent the input batch, where \( B \) is the batch size and \( F \) is the number of stacked frames, each flattened from an 84x84 grayscale image.

\textbf{Patch Embedding:} Given our input tensor \( X \in \mathbb{R}^{B \times (F \cdot 84 \cdot 84)} \), where:

The input is transformed into a sequence of embedded patches $X_p \in \mathbb{R}^{B \times (F \cdot N) \times E}$, where $N = \left(\frac{H}{p}\right)\left(\frac{W}{p}\right)$ is the number of patches per frame. This transformation involves unfolding the input into patches, linearly projecting each patch to an embedding space, and reshaping to match the Transformer's input format.

where \(W_{\text{proj}} \in \mathbb{R}^{(p^2) \times E}\) and \(b_{\text{proj}}\) represent the weights and bias of the linear projection layer, respectively. The resulting tensor, \(x_{\text{final}}\), with dimensions \([B, F, N, E]\), effectively encodes each patch into an embedding vector of size \(E\), ready for processing by the Transformer encoder; this process is heavily inspired by Vision Transformers (Doistevsky et al., 2021)\cite{dosovitskiy2021image}.

\textbf{Transformer Encoder and Action Value Prediction:} The Transformer encoder with gating mechanisms processes \(X_{\text{pos}}\), yielding \(X_{\text{enc}}\), which is then flattened and passed through fully connected layers to predict the Q-values for each action. The inclusion of gating mechanisms allows the DTQN model to effectively leverage the Transformer's capabilities for sequence processing while maintaining stability and enhancing learning, enabling it to learn complex policies based on sequences of environmental frames.

The gated Transformer encoder processes the sequence \(X_{\text{pos}}\), applying self-attention with gating and feed-forward with gating at each layer, yielding \(X_{\text{enc}}\):

\[
X_{\text{enc}} = \text{GatedTransformerEncoder}(X_{\text{pos}})
\]

where the \(\text{GatedTransformerEncoder}\) function represents the application of multiple layers of the GatedTransformerXLLayer, each consisting of a self-attention mechanism with a gating mechanism followed by a feed-forward network with another gating mechanism. Gating mechanisms are implemented using linear layers that combine the inputs and outputs of the self-attention and feed-forward sublayers, respectively, allowing for controlled information flow and stabilization of the learning process\cite{parisotto2019stabilizing}. 

Our

\textbf{Fully Connected Layers:}
Finally, \( X_{\text{enc}} \) is flattened and passed through fully connected layers to predict the Q-values for each action:
\begin{align*}
X_{\text{flat}} &= X_{\text{enc}}.\text{view}(B, -1) \\
X_{\text{fc1}} &= \text{ReLU}(W_{\text{fc1}} X_{\text{flat}} + b_{\text{fc1}}) \\
X_{\text{fc2}} &= \text{ReLU}(W_{\text{fc2}} X_{\text{fc1}} + b_{\text{fc2}}) \\
X_{\text{fc3}} &= \text{ReLU}(W_{\text{fc3}} X_{\text{fc2}} + b_{\text{fc3}}) \\
Q &= W_{\text{out}} X_{\text{fc3}} + b_{\text{out}}
\end{align*}
where \( W_{\text{fc1}}, W_{\text{fc2}}, W_{\text{fc3}}, W_{\text{out}} \) and \( b_{\text{fc1}}, b_{\text{fc2}}, b_{\text{fc3}}, b_{\text{out}} \) are the weights and biases of the respective fully connected layers, and \( Q \in \mathbb{R}^{B \times A} \) represents the predicted Q-values for \( A \) actions.
\subsubsection{Efficient Attention Model}
The Efficient Attention Model processes the input using linear projections, positional embeddings, a Transformer encoder, and attention pooling to output action values. The model architecture is formalized as follows:

\textbf{Input Projection:}
Let \( X \in \mathbb{R}^{B \times D} \) be the input batch, where \( B \) is the batch size and \( D \) is the input dimension. The input is first projected to a higher-dimensional embedding space:
\[
X_{proj} = XW_{proj} + b_{proj}
\]
where \( W_{proj} \in \mathbb{R}^{D \times E} \) and \( b_{proj} \in \mathbb{R}^{E} \) are the projection weights and biases, respectively, and \( E \) is the embedding size.

{Positional Embeddings:}
Positional embeddings are added to the projected input to retain positional information:
\[
X_{pos} = X_{proj} + P
\]
where \( P \in \mathbb{R}^{1 \times E} \) represents the positional embeddings, replicated for each element in the batch.

\textbf{GatedTransformerEncoder:}
The Transformer encoder processes the input with self-attention mechanisms:
\[
X_{enc} = \text{GatedTransformerEncoder}(X_{pos})
\]
where \(\text{gatedTransformerEncoder}\) includes multiple layers of multi-headed self-attention and position-wise feedforward networks.

\textbf{Attention Pooling:}
The output of the Transformer encoder is pooled using an attention mechanism to produce a context vector:
\[
c = \text{Attention Pooling}(X_{enc})
\]
where \( c \in \mathbb{R}^{B \times E} \) represents the pooled contextual information relevant for the decision-making process.

\textbf{Output Layer:}
Finally, the context vector is passed through fully connected layers to predict the action values:
\[
Q = cW_{out} + b_{out}
\]
where \( W_{out} \in \mathbb{R}^{E \times A} \) and \( b_{out} \in \mathbb{R}^{A} \) are the weights and biases of the output layer, and \( A \) is the number of possible actions.

\subsection{Training}
Let \( \mathcal{M} \) represent the memory buffer of size \( N \), where an experience \( e_t = (s_t, a_t, r_{t+1}, s_{t+1}) \) is stored at time \( t \). Sampling a minibatch of size \( k \) from \( \mathcal{M} \) can be mathematically represented by \( B = \{e_1, e_2, ..., e_k\} \subseteq \mathcal{M} \).

For the \(\epsilon\)-greedy strategy, the action \(a_t\) at time \(t\) is chosen as follows:
\[
a_t = 
\begin{cases} 
\text{random action from } \mathcal{A}', & \text{with probability } \epsilon \\
\arg\max_{a \in \mathcal{A}'} Q(s_t, a), & \text{with probability } 1 - \epsilon
\end{cases}
\]
where \(\epsilon\) is the exploration rate, and \(Q(s_t, a)\) represents the action-value function estimate for state \(s_t\) and action \(a\).

The epsilon decay process is mathematically represented as:
\[
\epsilon \leftarrow \max(\epsilon_{\text{end}}, \epsilon \cdot \text{decay\_rate})
\]
where \(\text{decay\_rate}\) is a factor less than 1 used to decrease \(\epsilon\) after each episode.

\subsubsection{Q-value Update}
Following the selection of an action \(a_t\), the Q-values are updated using the Bellman equation:
\begin{align}
Q(s_t, a_t) &\leftarrow Q(s_t, a_t) + \alpha \left( r_{t+1} + \gamma \max_{a'} Q(s_{t+1}, a') \right. \nonumber \\
            &\quad \left. - Q(s_t, a_t) \right) \label{eq:update} \\
            &= Q(s_t, a_t) + \alpha \Delta Q \label{eq:simple_update}
\end{align}

where \(\alpha\) is the learning rate, \(\gamma\) is the discount factor, \(a'\) represents all possible actions from the state \(s_{t+1}\), and \(\Delta Q\) is the temporal-difference error.

\subsubsection{Loss Function Selection}
The loss function used for training is selected dynamically based on the empirical observation of the magnitude of the loss values. We carefully observe the behavior of the loss over a significant number of episodes to determine if the training should be restarted with a different loss function. The criteria for selecting the loss function are based on the observed behavior as follows:
\begin{itemize}
  
\item Observing Flat Loss: If the loss remains flat and does not decrease or increase over a prolonged period of time, it suggests that the current loss function might be too robust to outliers causing the Q-Network to underfit and struggle to determine the distance between the current action and the best action. For this scenario we switch from Huber to MSE loss
\item High Volatility: If using MSE results in highly volatile loss measurements. If the MSE loss is too volatile a NaN may occur, therefore in this scenario we switch to Huber loss.

\end{itemize}

\subsection{Experimental Setup}
The training of both DCQN and DTQN models follows a similar protocol, with specific adaptations to each model's architecture This section will cover all the different configurations for each of the 6 individually trained Agents.

\subsubsection{Hyperparameters}

\begin{table}[h]
\centering
\caption{Training Hyperparameters for Different Agents}
\label{tab:hyperparameters}
\begin{tabular}{@{}>{\raggedright\arraybackslash}p{0.22\linewidth} 
                 >{\raggedright\arraybackslash}p{0.10\linewidth} 
                 >{\centering\arraybackslash}p{0.10\linewidth} 
                 >{\centering\arraybackslash}p{0.10\linewidth} 
                 >{\centering\arraybackslash}p{0.08\linewidth} 
                 >{\centering\arraybackslash}p{0.13\linewidth} 
                 >{\raggedright\arraybackslash}p{0.13\linewidth}@{}}
\toprule
\textbf{Model} & \textbf{Game} & \textbf{Learning Rate} & \textbf{Discount Factor (\(\gamma\))} & \textbf{Batch Size} & \textbf{Replay Buffer Size} & \textbf{Target Network Update} \\ 
\midrule
DCQN & Centipede & \(1 \times 10^{-4}\) & 0.99 & 32 & \(1 \times 10^6\) & Every 500 steps \\
DTQN & Centipede & \(2 \times 10^{-4}\) & 0.99 & 32 & \(1 \times 10^6\) & Every 500 steps \\
DCQN  & Asteroids & \(1 \times 10^{-4}\) & 0.99 & 32 & \(1 \times 10^6\) & Every 100 steps \\
DTQN & Asteroids & \(3 \times 10^{-4}\) & 0.99 & 32 & \(1 \times 10^6\) & Every 100 steps \\
DCQN & Space Invaders & \(2 \times 10^{-4}\) & 0.99 & 32 & \(1 \times 10^6\) & Every 500 steps \\
DTQN & Space Invaders & \(1 \times 10^{-4}\) & 0.99 & 32 & \(1 \times 10^6\) & Every 500 steps \\
\bottomrule
\end{tabular}
\end{table}

\subsubsection{Optimizer}
All six of the Agents are trained using the AdamW Optimizer, with the learning rates and weight decay's specified above.

\subsubsection{Training Duration}
Each Agent is trained for 10,000 episodes, this is a varying number of frames based on the agent's performance as well as the game environment.
\subsubsection{Evaluation Protocol}
Model evaluation is conducted at regular intervals during training to assess performance and convergence:

\subsubsection{Evaluation Frequency}
Both models are evaluated every 500 episodes, providing periodic snapshots of their learning progress and capabilities. This done through performing a fully deterministic eval.

\subsubsection{Evaluation Metrics}
Performance is quantified using the average rewards per episode over 5 evaluation episodes, providing a robust measure of the models' effectiveness.

\subsubsection{Exploration Strategy}
During training, the agent employs an $\epsilon$-greedy policy where $\epsilon$ starts at 1.0 and exponentially decays per episode towards a minimum of 0.1, according to:
\[
\epsilon \leftarrow \max(\epsilon_{\text{end}}, \epsilon \cdot \epsilon_{\text{decay}})
\]
with $\epsilon_{\text{end}} = 0.1$ and the decay factor computed as:
\[
\epsilon_{\text{decay}} = \left(\frac{\epsilon_{\text{end}}}{\epsilon_{\text{start}}}\right)^{\frac{1}{\text{n\_episodes}}}
\]
During evaluation, no epsilon is used to evaluate the model using pure deterministic actions.

\subsubsection{Implementation Details}
Both models are implemented using PyTorch. The training and evaluation procedures are facilitated by a custom training loop that handles interactions with the environment, model updates, and logging:

\subsubsection{Environment Interaction}
We use the OpenAI Gym interface to interact with Atari environments, with each action chosen by the model affecting the state observed in subsequent steps.

\subsubsection{Model Updates}
The model parameters are updated using batches of experiences sampled from the replay buffer, with the loss calculated as described in the loss function subsection.

\subsubsection{Logging and Monitoring}
Training progress is monitored using TensorBoard, where key metrics such as average return, loss, and \(\epsilon\) values are logged for visualization and analysis.

\section{Results}
In this section, we will discuss the results of our RL Agents after being trained over 10000 episodes, in this section we address not only the results of the evaluations but also the observations from training.
\begin{figure}[ht]
    \centering
    \begin{minipage}[b]{0.22\textwidth}
        \centering
        \includegraphics[width=\textwidth]{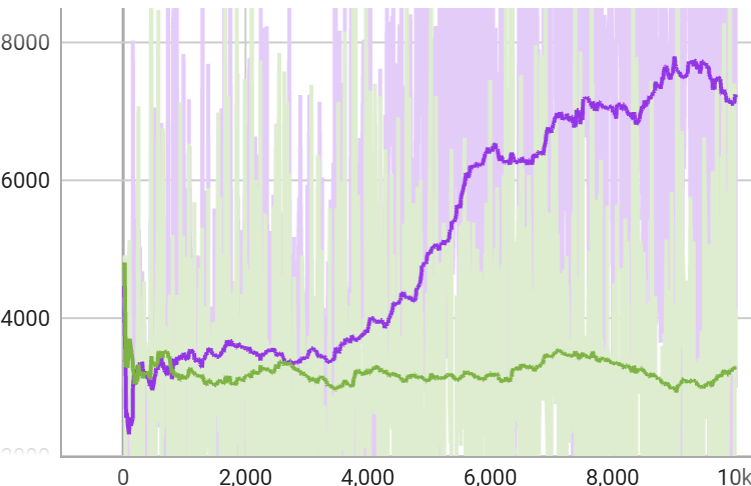}
        \caption{Centipede Reward over 10k episodes. DTQN is depicted in Green and DCQN is depicted in Purple}
        \label{fig:figure4}
    \end{minipage}
    \hfill 
    \begin{minipage}[b]{0.22\textwidth}
        \centering
        \includegraphics[width=\textwidth]{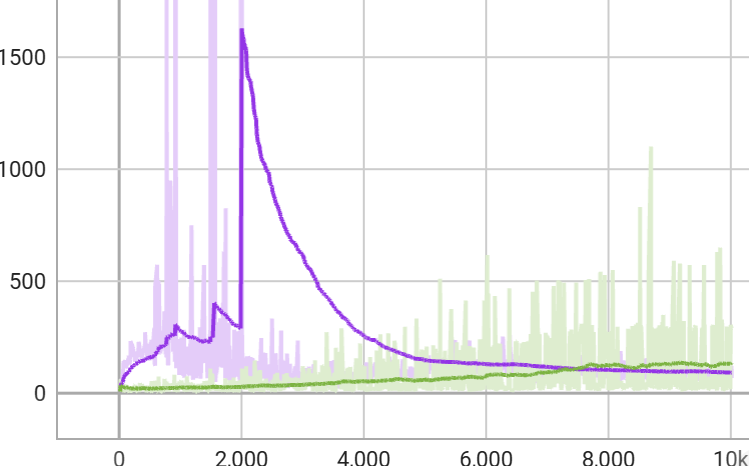}
        \caption{Centipede Loss over 10k Episodes DTQN is depicted in green and DCQN is depicted in purple}
        \label{fig:figure5}
    \end{minipage}
    \hfill 
    \begin{minipage}[b]{0.22\textwidth}
        \centering
        \includegraphics[width=\textwidth]{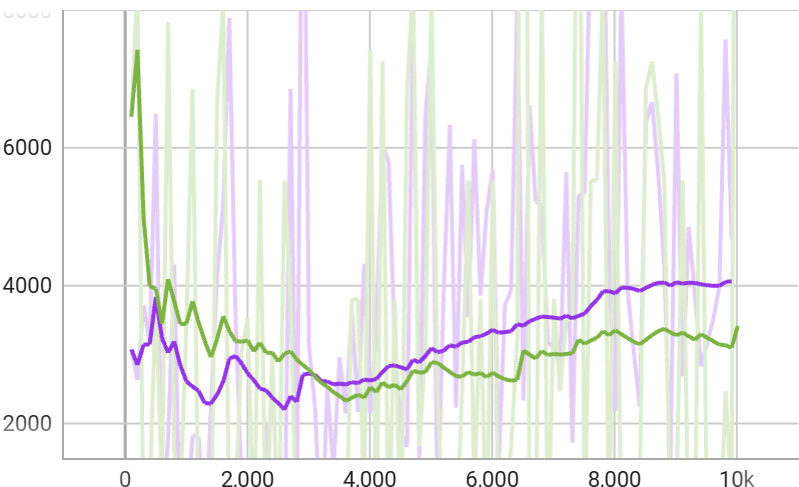}
        \caption{Centipede Average Eval Reward over 10k episodes, evaluations are performed every 500 steps. DTQN is depicted in green DCQN is depicted in purple.}
        \label{fig:figure6}
    \end{minipage}
    \hfill 
    \begin{minipage}[b]{0.22\textwidth}
        \centering
        \includegraphics[width=\textwidth]{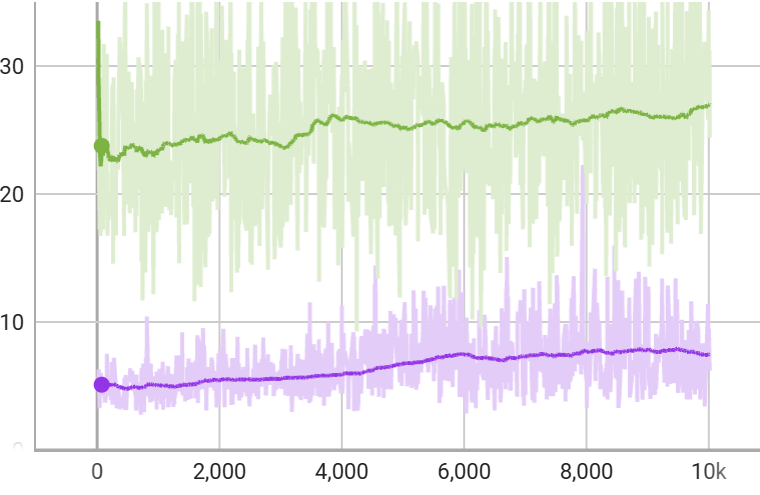}
        \caption{Deep Convolutional Q-Network (DCQN) is illustrated using a purple color scheme, while the Deep Transformer Q-Network (DTQN) is depicted in green. Environment time over 10k episodes.}
        \label{fig:Traditional Q-Network Implementation}
    \end{minipage}
\end{figure}
\begin{figure}[ht]
    \centering
    \begin{minipage}[b]{0.22\textwidth}
        \centering
        \includegraphics[width=\textwidth]{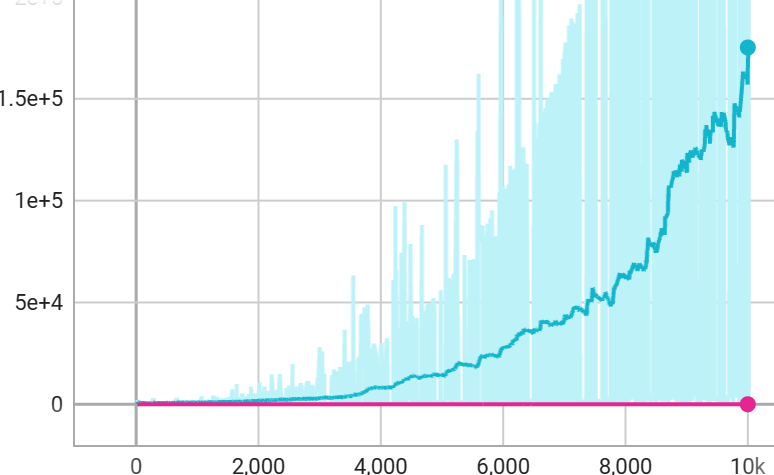}
        \caption{Asteroids Loss over 10k episodes. DTQN is depicted in green and DCQN is depicted in purple.}
        \label{fig:asteroids_loss}
    \end{minipage}
    \hfill 
    \begin{minipage}[b]{0.22\textwidth}
        \centering
        \includegraphics[width=\textwidth]{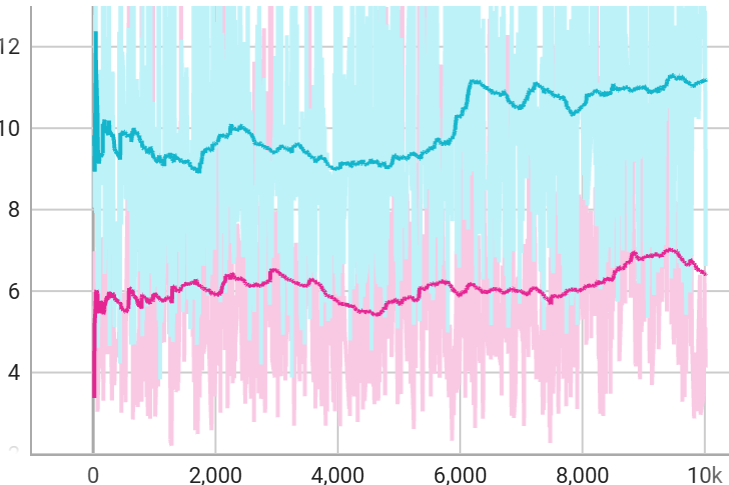}
        \caption{Environment time for Asteroids over 10k episodes. DTQN is depicted in green and DCQN in purple.}
        \label{fig:asteroids_env_time}
    \end{minipage}
    \hfill 
    \begin{minipage}[b]{0.22\textwidth}
        \centering
        \includegraphics[width=\textwidth]{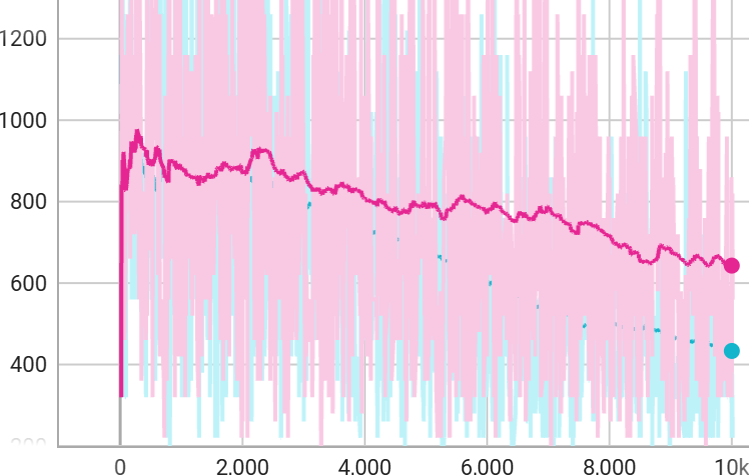}
        \caption{Total reward for Asteroids over 10k episodes. DTQN in green and DCQN in purple.}
        \label{fig:asteroids_reward}
    \end{minipage}
    \hfill 
    \begin{minipage}[b]{0.22\textwidth}
        \centering
        \includegraphics[width=\textwidth]{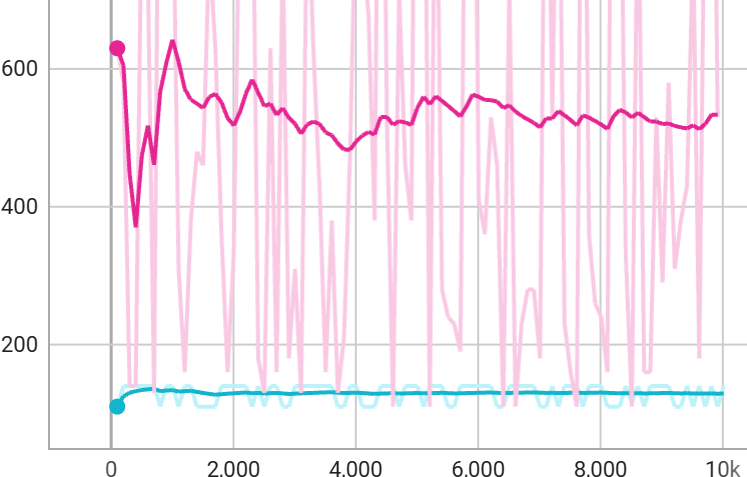}
        \caption{Average evaluation reward for Asteroids, evaluated every 500 steps over 10k episodes. DTQN is shown in green, DCQN in purple.}
        \label{fig:asteroids_avg_reward}
    \end{minipage}
\end{figure}

\begin{table*}[ht]
\centering
\caption{Summary of Game Performance}
\label{tab:game_performance}
\begin{tabular}{@{}lcccc@{}}
\toprule
Game & Model & Lowest Reward & Highest Reward & Average Reward \\ 
\midrule
Centipede & DCON & 7714 & 13853 & 10152.4 \\
Centipede & DTQN & 3505 & 27908 & 13266.2 \\
Asteroids & DCON & 1300 & 680 & 920 \\
Asteroids & DTQN & 140 & 140 & 140 \\
Space Invaders & DTQN & 285 & 285 & 285 \\
Space Invaders & DCQN & 245 & 535 & 411.7 \\
\bottomrule
\end{tabular}
\end{table*}
We see that the Convolutional architecture outperforms the Transformer architectures across all games but Centipede in evaluation. However, as evidenced from \ref{fig:figure3} This may be an outlier since during training the DCQN trended higher towards the later episodes. All version of the DTQN were slower than the DCQN except for Asteroids, in evaluation this is because the Agent is quickly dying and is not related to computational efficiency. Without pretraining the Transformer is able to detect objects directly overhead the player, this results in high performance playing centipede, as well as shooting straight upwards in Space Invaders. In Centipede, in which both models utilize Huber Loss, we see that the Transformer is flat across all benchmarks for the entire duration of training, and we notice significant increase in the reward of the DCQN after the loss function spikes and then decreases, we can assume that these models have reached a plateau for traditional Q-Learning based on the behavior and these measures.
\subsection{Model Behavior}
\subsubsection{Centipede}
The DTQN Agent learned quickly in Centipede that an easy way to maximize points was to sit in one area the whole time and continuously shoot. Since the Centipede is guaranteed to move side to side across the screen shooting quickly upwards allowed for large sections and points to be accumulated at one time. Not moving is also an advantage because the Spider is more likely to cross the firing path than it is for the Agent to collide with it, resulting in more chances to reap rewards. This allows it to outperform the DCQN over 10000 episodes in this game. The DCQN Agent learns to dodge occasionally, performing best when the Centipede consists of a larger number of pixels (less of it has been destroyed). However, the DCQN Agent also shows signs of the same camping strategy.
\subsubsection{Space Invaders}There is a significant difference in performance between the two agents visually in Space Invaders, the DTQN does not move at any point of time, this is problematic because the Agent is stuck at 285 points, perhaps using a very low epsilon would allow for the Agent to perform better during the deterministic stages. The DCQN learns slightly how to dodge towards the end of the episodes which allows for it to perform on average much better than the DTQN.
\subsubsection{Asteroids} The DCQN learns the process normally, learning to shoot the moving objectts, the DTQN has not learned anything over 10000 episodes even with the loss significantly increasing towards the end. We believe we can attribute this to the feature loss during linear projection. We also believe that this Agent could perform better if we established a penalty on repeated actions.
\section{Discussion}
This study presents a comparative analysis of Deep Convolutional Q-Networks (DCQN) and Deep Transformer Q-Networks (DTQN) across three different Atari games, highlighting the strengths and limitations of each model in the context of reinforcement learning. The findings reveal significant differences in performance and learning behaviors between the two architectures, offering insights into their respective efficiencies and strategic capabilities.

The results indicate that DCQN generally outperforms DTQN in terms of both speed and average reward across most games, with the notable exception of Centipede, where DTQN demonstrates superior performance. This exception can be attributed to DTQN's ability to capitalize on specific game dynamics, such as the predictable movement patterns of the Centipede, which align well with the model's architectural strengths in handling sequential data.

However, the DTQN's performance in other games, particularly Asteroids, was underwhelming. The lack of learning progress in Asteroids suggests that DTQN may suffer from significant feature loss during the linear projection phase, which inhibits its ability to effectively process and respond to dynamic game environments. This feature loss might be mitigated by adjusting the model's architecture or training protocol, such as introducing penalties for repeated actions to encourage more diverse strategic exploration.

Moreover, the DCQN's ability to adapt and improve performance over time, as seen in Space Invaders, underscores its utility in scenarios where agents must learn to navigate and respond to evolving threats dynamically. The contrast in performance between DCQN and DTQN in this game also highlights the potential challenges DTQN faces in environments requiring rapid and complex spatial decision-making.

These observations suggest that while Transformer-based models like DTQN hold promise for enhancing sequential decision-making in games, their current implementations may need further refinement to fully exploit their capabilities across a broader range of scenarios. Meanwhile, DCQN remains a robust and reliable choice for many typical game environments, balancing performance with computational efficiency. For the most part this refinement involves pre-training or simply using larger models.

\subsection{Limitations}
\subsubsection{Patch Embedding Size}
Our Transformer implementation uses the same 16x16 patch size as ViT[]. However, in the original paper, the 16x16 patch was designed for 224x224 images; in this study, we use a 16x16 patch on an 84x84 image which is not proportional segmentation of information. Additionally, most implementations like Deit and other lightweight models rely on ViT Feature Extractor to prepossess the images since it would be computationally intensive to do so inside of the model the reason we do not use a 6x6 patch size, which would line up more with ViT in terms of proportionality is that it increases the param size by approximately 700 percent.
\subsubsection{Model Size}
In our study we only tested the DQN and DTQN models in the 36M-39M parameter range, it should be considered that one model might perform differently with a different level of complexity even with the same architecture.In other studies 
\subsubsection{Model Architecture}
Our proposed DTQN is one of many possible implementations of Transformers in RL, based on the existing literature, different implementations lead to different findings. For our specific architecture replacing the patch embedding and using Convolutions to extract features from the model before applying the positional embedding leads to a 300 percent reduction in the number of parameters. As a result of this the Embed Size of the Transformer can be doubled to get back to our 30 million parameter class model.


\begin{appendices}
\section{Sequential Replay Buffer}
The Sequential Replay Buffer is a modified version of the standard experience replay buffer, designed to handle sequences of observations, actions, rewards, and terminal states. This buffer is particularly useful for training agents that rely on temporal information, such as recurrent neural networks or agents that operate on sequences of observations.

The key components of the Sequential Replay Buffer are as follows:

\subsection{Initialization}
The Sequential Replay Buffer is initialized with the following parameters:
\begin{align*}
    \mathcal{C} &= \text{capacity of the replay buffer} \\
    \mathcal{S} &= \text{shape of the state/observation} \\
    \mathcal{B} &= \text{batch size} \\
    \mathcal{L} &= \text{sequence length}
\end{align*}

The buffer is implemented using a \textit{LazyMemmapStorage} to efficiently store the experiences, and a \textit{TensorDictReplayBuffer} to manage the sampling and retrieval of experiences.

\subsection{Adding Experiences}
To add a sequence of experiences to the buffer, the \texttt{add} method is used:
\begin{align*}
    \text{add}(\mathbf{s}, \mathbf{a}, \mathbf{r}, \mathbf{s'}, \mathbf{d})
\end{align*}
where:
\begin{align*}
    \mathbf{s} &= \text{sequence of states/observations} \in \mathbb{R}^{\mathcal{L} \times \mathcal{S}} \\
    \mathbf{a} &= \text{sequence of actions} \in \mathbb{Z}^{\mathcal{L} \times 1} \\
    \mathbf{r} &= \text{sequence of rewards} \in \mathbb{R}^{\mathcal{L} \times 1} \\
    \mathbf{s'} &= \text{sequence of next states/observations} \in \mathbb{R}^{\mathcal{L} \times \mathcal{S}} \\
    \mathbf{d} &= \text{sequence of terminal states} \in \{0, 1\}^{\mathcal{L} \times 1}
\end{align*}

The method ensures that the input tensors have the correct shape and stores the sequence in the replay buffer.

\subsection{Sampling Experiences}
To sample a batch of sequences from the replay buffer, the \texttt{sample} method is used:
\begin{align*}
    \text{sample}() \rightarrow \{\mathbf{s}, \mathbf{a}, \mathbf{r}, \mathbf{s'}, \mathbf{d}\}
\end{align*}
The method returns a dictionary containing the sampled sequences of states, actions, rewards, next states, and terminal states.

\subsection{Implementation Details}
The Sequential Replay Buffer is implemented using the \textit{LazyMemmapStorage} and \textit{TensorDictReplayBuffer} from the \texttt{torchrl} library. The storage keys are initialized with the appropriate shapes to accommodate the sequence-level data. The \texttt{add} and \texttt{sample} methods are implemented to handle the sequence-level data accordingly.
\begin{verbatim}

\end{verbatim}
\section{Convolutional Transformer Model}
The Convolutional Transformer Model integrates convolutional features into a Transformer architecture. The model can be expressed as follows:

\paragraph{Convolutional Feature Extraction:}
Let \( I \in \mathbb{R}^{B \times C \times H \times W} \) represent the input image batch. The convolutional layers extract features:
\[
F = \text{ConvLayers}(I)
\]
where \( F \in \mathbb{R}^{B \times F_{dim} \times H' \times W'} \) and \( F_{dim} \) is the number of feature dimensions after the convolutional layers.

\paragraph{Feature Flattening and Positional Embedding:}
The features are flattened and combined with positional embeddings:
\[
F_{flat} = F.\text{reshape}(B, F_{dim} \times H' \times W') + P
\]
where \( P \in \mathbb{R}^{1 \times (F_{dim} \times H' \times W')} \) are the positional embeddings.

\paragraph{Transformer Encoder:}
The flattened features are then processed by a Transformer encoder:
\[
F_{enc} = \text{Gated Transformer XL}(F_{flat})
\]
similar to the Efficient Attention Model.

\paragraph{Output Prediction:}
The output of the Transformer encoder is passed through a final fully connected layer to predict the action values:
\[
Q = F_{enc}W_{out} + b_{out}
\]
where \( W_{out} \in \mathbb{R}^{(F_{dim} \times H' \times W') \times A} \) and \( b_{out} \in \mathbb{R}^{A} \).
\section{Gated Transformer-XL Layer Mathematical Description}
The Gated Transformer-XL architecture enhances the standard Transformer by integrating gating mechanisms at each layer to control the flow of information and improve learning dynamics. The detailed operations within a Gated Transformer-XL layer are as follows:

\subsection{Layer Components}
Each GatedTransformerXLLayer comprises the following components:
\begin{itemize}
    \item A multi-head self-attention mechanism.
    \item Two feed-forward neural networks.
    \item Layer normalization applied before each sub-layer.
    \item Dropout for regularization.
    \item Gating mechanisms to combine inputs from different sub-layers.
\end{itemize}

\subsection{Mathematical Formulation}
Given an input sequence $X \in \mathbb{R}^{n \times d}$, where $n$ is the sequence length and $d$ is the model dimension, the operations are defined as follows:

\subsubsection{Self-Attention with Gating}
The self-attention mechanism computes the output as:
\[
    \text{Attn}(Q, K, V) = \text{softmax}\left(\frac{QK^T}{\sqrt{d_k}}\right)V
\]
where $Q$, $K$, and $V$ are the queries, keys, and values respectively, all derived from the layer input $X$. The output of the attention layer is then gated as follows:
\[
    G_1 = \sigma(W_g[X, \text{Attn}(Q, K, V)] + b_g)
\]
where $W_g \in \mathbb{R}^{2d \times d}$ and $b_g \in \mathbb{R}^d$ are the gating parameters, and $\sigma$ is the sigmoid activation function. The final output after the gating and residual connection is:
\[
    Y = X + \text{Dropout}(G_1)
\]

\subsubsection{Feed-Forward Network with Gating}
The feed-forward network consists of two linear transformations with a ReLU activation in between:
\[
    F(X) = \text{ReLU}(W_1X + b_1)W_2 + b_2
\]
where $W_1 \in \mathbb{R}^{d \times f}$, $b_1 \in \mathbb{R}^f$, $W_2 \in \mathbb{R}^{f \times d}$, and $b_2 \in \mathbb{R}^d$ are the network parameters, and $f$ is the dimensionality of the feed-forward layer. The output of the feed-forward layer is gated:
\[
    G_2 = \sigma(W_g'[Y, F(Y)] + b_g')
\]
where $W_g' \in \mathbb{R}^{2d \times d}$ and $b_g' \in \mathbb{R}^d$ are additional gating parameters. The final output from the feed-forward network is:
\[
    Z = Y + \text{Dropout}(G_2)
\]

The output $Z$ of each layer is then normalized:
\[
    \text{Output} = \text{LayerNorm}(Z)
\]

\subsection{Layer Stacking and Model Output}
Multiple GatedTransformerXLLayers are stacked to form the full GatedTransformerXL model. The output of each layer is fed as the input to the next layer. The final layer output is used for downstream tasks such as classification or regression.
\end{appendices}
\end{document}